\documentclass{article}

\usepackage[dblblindworkshop, final]{neurips_2025}

\usepackage{amsfonts}       
\usepackage{nicefrac}       
\usepackage{microtype}      
\usepackage{xcolor}         
\usepackage{multirow}
\usepackage{natbib}
\usepackage{arydshln} 
\usepackage{subcaption}
\usepackage{makecell} 
\usepackage{placeins} 
\usepackage{graphicx}       
\usepackage{subcaption} 
\usepackage[bottom]{footmisc}
\usepackage{subcaption}
\usepackage{bbding}
\usepackage{amsmath}
\usepackage{soul}
\usepackage{hyperref} 
\usepackage{tablefootnote} 
\usepackage{caption} 
\usepackage{ulem}
\usepackage{cancel}
\usepackage{times}
\usepackage{xcolor}
\usepackage{latexsym}
\usepackage{amssymb}
\usepackage[T1]{fontenc}
\usepackage{booktabs}
\usepackage{wrapfig}
\usepackage{tabularx}
\usepackage[utf8]{inputenc}
\usepackage{tcolorbox}
\usepackage{microtype}
\usepackage{colortbl} 
\usepackage{inconsolata}
\usepackage{graphicx}
\usepackage{enumitem}
\usepackage{url}

\workshoptitle{Efficient Reasoning}

\title{The Zero-Step Thinking: An Empirical Study of Mode Selection as  Harder Early Exit in Reasoning Models}

\author{Yuqiao Tan$^{1,2}$, Shizhu He$^{1,2}$\thanks{Corresponding author} ,  \textbf{Kang Liu}$^{1,2,3}$, \textbf{Jun Zhao}$^{1,2}$  \\
    $^1$ The Key Laboratory of Cognition and Decision Intelligence for Complex Systems, \\
    Institute of Automation, Chinese Academy of Sciences, Beijing, China \\
    $^2$ School of Artificial Intelligence, University of Chinese Academy of Sciences, Beijing, China \\
    $^3$ Shanghai Artificial Intelligence Laboratory \\
  {tanyuqiao2025@ia.ac.cn} {\{shizhu.he, jzhao, kliu\}@nlpr.ia.ac.cn}\\}

\begin{document}

\maketitle

\begin{abstract}
Reasoning models have demonstrated exceptional performance in tasks such as mathematics and logical reasoning, primarily due to their ability to engage in step-by-step thinking during the reasoning process. However, this often leads to overthinking, resulting in unnecessary computational overhead. To address this issue, Mode Selection aims to automatically decide between Long-CoT (Chain-of-Thought) or Short-CoT by utilizing either a \textsc{Thinking} or \textsc{NoThinking} mode. Simultaneously, Early Exit determines the optimal stopping point during the iterative reasoning process. Both methods seek to reduce the computational burden.
In this paper, we first identify Mode Selection as a more challenging variant of the Early Exit problem, as they share similar objectives but differ in decision timing. While Early Exit focuses on determining the best stopping point for concise reasoning at inference time, Mode Selection must make this decision at the beginning of the reasoning process, relying on pre-defined fake thoughts without engaging in an explicit reasoning process, referred to as zero-step thinking.
Through empirical studies on nine baselines, we observe that prompt-based approaches often fail due to their limited classification capabilities when provided with minimal hand-crafted information. In contrast, approaches that leverage internal information generally perform better across most scenarios but still exhibit issues with stability.
Our findings indicate that existing methods relying solely on the information provided by models are insufficient for effectively addressing Mode Selection in scenarios with limited information, highlighting the ongoing challenges of this task. Our code is available at \url{https://github.com/Trae1ounG/Zero_Step_Thinking}.
\end{abstract}

\begin{figure}[ht]
  \begin{center}
  {\includegraphics[width=0.9\textwidth]{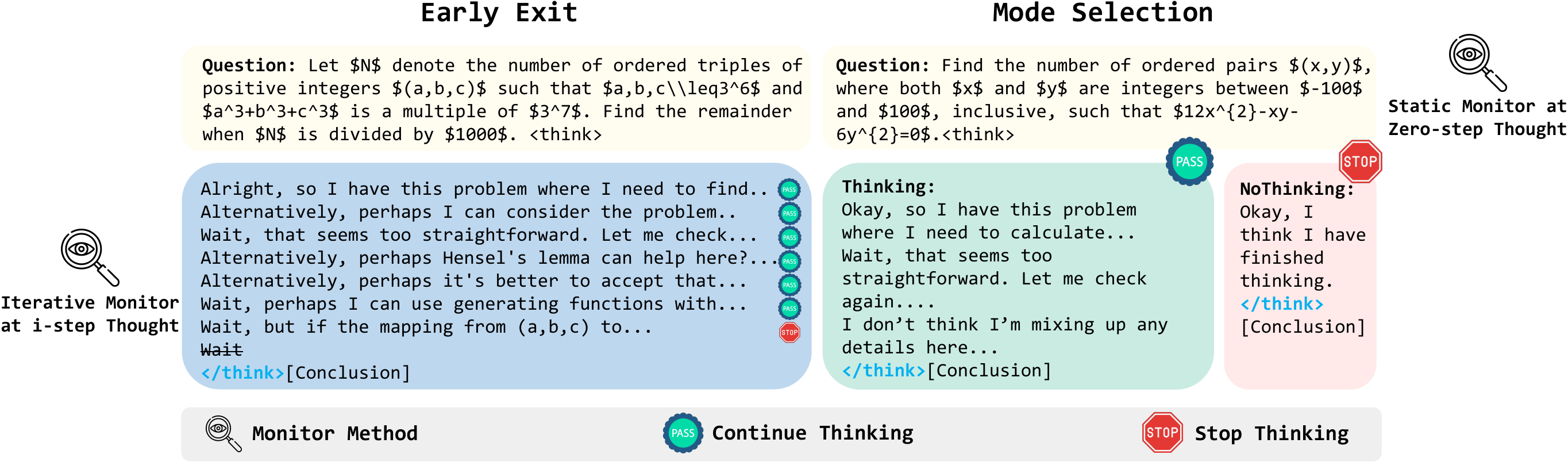}}
  \caption{Illustration of \textbf{Early Exit} and \textbf{Mode Selection}. Early Exit employs an iterative monitor to decide whether to stop reasoning at the end of each step in the thought process. In contrast, Mode Selection operates prior to explicit reasoning, determining the optimal thinking strategy at zero-step. While both methods can leverage the same monitoring mechanisms, they differ in their timing: Early Exit monitors iteratively during the reasoning process, whereas Mode Selection monitors only once at the beginning to append \texttt{</think>} and terminate the reasoning process.}
  \label{fig:intro}
  \end{center}
  
\end{figure} 

\section{Introduction}
Recent advances in large reasoning models (LRMs), such as DeepSeek-R1~\citep{guo2025deepseek}, OpenAI o1~\citep{jaech2024openai}, QwQ~\citep{qwen-qwq}, and others~\citep{gemini2.5-pro, Claude3S, sky-t1}, have demonstrated significant progress in complex reasoning capabilities by increasing inference-time compute~\citep{muennighoff2025s1}. These models achieve success through long chain-of-thought (CoT)\citep{wei2022chain} processes, which involve behaviors such as exploration, self-reflection, and verification\citep{li2025system, gandhi2025cognitive, ghosal2025does}. However, this strength can also become a limitation. Previous studies have shown that reasoning models often tend to overthink, exploring additional reasoning paths even after arriving at a correct answer~\citep{chen2024not,sui2025stop,wang2025adaptive}. 

The adaptive nature of human cognition commonly switches between System 1 and System 2 based on task difficulty.  Inspired by this, recent research on \textbf{Adaptive Thinking} seeks to enable a single, powerful reasoning model to operate in two distinct modes: its native long-CoT mode for complex problems and an efficient short-CoT mode for simpler tasks~\citep{zhang2025long, tu2025learning, liu2025qfft}. Many of these approaches utilize reinforcement learning to train reasoning models to think more concisely and switch modes adaptively~\citep{zhang2025adaptthink, fang2025thinkless, luo2025ada}.

Another efficient reasoning solution is \textbf{Early Exit}, which aims to determine the optimal stopping point in the reasoning process by truncating it to directly produce an answer~\citep{yang2025dynamic, jiang2025flashthink, zhang2025reasoning}. This approach helps LRMs avoid redundant reasoning paths that may lead to overthinking and potentially performance degradation. Existing studies have demonstrated that reasoning models can inherently perform short-CoT in \textsc{NoThinking} mode by appending fake thoughts directly (e.g., \texttt{<think>} Okay, I think I have finished thinking. \texttt{</think>}), thereby significantly reducing token usage while maintaining strong performance~\citep{ma2025reasoning,li2025thinkless,li2025dynamicmind, zhu2025can}. 

In this study, we focus on a specific subtype of Adaptive Thinking called \textbf{Mode Selection}, which determines the preferred prompt mode without relying on an explicit reasoning trajectory. This process can be regarded as zero-step thinking~\citep{liang2025thinkswitcher}.
Interestingly, we find that Mode Selection is conceptually similar to Early Exit but differs in decision timing and is significantly more challenging. As shown in Figure~\ref{fig:intro}, Early Exit operates iteratively by taking previously generated thinking tokens into account to make decisions at the end of each reasoning step. This iterative process not only delivers better performance but also reduces token usage due to choosing optimal stop points~\citep{yang2025dynamic, zhang2025reasoning, gong2025efficient}. In contrast, Mode Selection relies solely on input tokens with human-crafted fake thoughts to determine the optimal mode at the outset of the reasoning process, making it a more difficult task due to the limited information available. Intuitively, both methods employ a monitoring mechanism $\text{Exit}(\cdot)$ to facilitate exit decisions. This observation raises our core research question: \textit{Can existing well-performing Early Exit methods effectively address the harder Mode Selection problem?}

To better understand this question, we conduct empirical studies using several methods from Early Exit, categorized into two distinct types: \textit{Prompt-based} and \textit{Internal States-based} methods, which utilize different prompt templates and internal model information for Mode Selection respectively. We rigorously evaluate these methods across diverse reasoning benchmarks, including GSM8K~\citep{cobbe2021training}, MATH-500~\citep{hendrycks2021measuring}, AIME25~\citep{AIME}, and GPQA-D~\citep{rein2024gpqa}, covering both mathematical and scientific reasoning tasks.
Our experiments reveal that prompt-based approaches often fail due to their limited classification capabilities when provided with minimal information. In contrast, internal states-based methods, which leverage internal model information, perform better in most scenarios but still struggle with stability.
Through the analysis of various metrics, including ROC-AUC, Expected Calibration Error (ECE)~\citep{naeini2015obtaining}, and Brier score~\citep{glenn1950verification}, we observe that existing evaluation metrics are insufficient for fully assessing and explaining the underlying reasons behind the performance of different methods. This highlights that Mode Selection is a highly challenging task, requiring deeper investigation into the model's internal mechanisms of \textsc{Thikning} and \textsc{NoThinking} to achieve effective solutions.
We summarize our contributions as follows:
\begin{itemize}[leftmargin=*]
    \item We first formalize the Mode Selection task as a more challenging variant of the Early Exit problem, which shares similar objectives but differs in decision timing. Both are unified under the same $\text{Exit}(\cdot)$ function.
    \item Through extensive experiments across various datasets, we observe that prompt-based approaches often fail due to their limited classification capabilities, whereas internal states-based methods generally perform better but still struggle with stability.
    \item By utilizing metrics such as ROC-AUC, ECE, and Brier score, we uncover the instability in method decisions and the model's inherent abilities, emphasizing the persistent challenges of this task.
\end{itemize}


\section{Preliminaries of Early Exit in LRMs}
\label{sec:early_exit}
\textit{Overthinking} is a critical phenomenon in large reasoning models (LRMs), where model performance initially improves with extended reasoning chains but deteriorates beyond a certain point~\citep{chen2024not,sui2025stop,wang2025adaptive}. This implies that thinking longer is not always better. Existing Early Exit methods aim to determine the optimal moment to terminate the reasoning process. In reasoning-focused LRMs, these models function as System 2, where the generation process is divided into two stages: \textit{thinking} and \textit{conclusion}, collectively referred to as \textsc{Thinking} mode~\citep{deepseekv3.1, jaech2024openai}:
\begin{equation}
    \textsc{Thinking}:  \underbrace{\text{[Prompt]} + \texttt{<think>}}_{\text{Input}} + \underbrace{\text{[Thoughts]} + \texttt{</think>} + \text{[Conclusion]}}_{\text{Generate}}
\end{equation}
Here, \texttt{<think>} and \texttt{</think>} serve as the delimiters for the beginning and end of the reasoning process, respectively. \text{[Thoughts]} represents the detailed reasoning process, while \text{[Conclusion]} contains the concise answer. Formally, the reasoning segment \text{[Thoughts]} reveals the trajectory of problem-solving, commonly referred to as Chain of Thought (CoT)~\citep{wei2022chain}.
At a micro level, temporally consecutive tokens in \text{[Thoughts]} can be grouped into short semantic, logical, and grammatically coherent chunks $T_j$:
\begin{equation}
    \texttt{[Thoughts]} = T \mathrel{\mathop:}= [T_1, \cdots, T_n]
\end{equation}
Each reasoning chunk in this paradigm is auto-regressively generated by conditioning on the question $Q$ and the preceding reasoning token: $T_i=\text{LRM}(Q,T_{<i})$. For simplicity, the Early Exit method performs $\text{Exit}(Q, T_{<i})$ based on the question and the reasoning trajectory up to $T_{<i}$. This determines whether to terminate the reasoning process at chunk $T_i$ by appending \texttt{</think>}.
The fundamental technique in Early Exit is predicting whether the current information is sufficient to solve the question. To this end, we collect several methods capable of expressing their confidence in the current reasoning process. Considering existing methods in LRMs, we categorize them into three types based on their information sources: \textit{Prompt-based} and \textit{Internal States-based}. Each type implements the $\text{Exit}(Q,T_{<i})$ function differently, leveraging distinct resources to determine when to stop reasoning.

\subsection{Prompt-based Early Exit}

\textbf{\textsc{FlashThink}}~\citep{jiang2025flashthink} employs a separate verification model, $\pi_{\phi}$, parameterized by $\phi$, along with a specific prompt. This model is used to decide when to stop reasoning by iteratively querying at the end of each thought:
\begin{equation}
    s_i = \pi_\phi(Q, T_{<i},\text{[Verification Prompt]})    
\end{equation}
where if $s_i$ is true will stop the following thinking process.

\textbf{\textsc{Prompt Confidence (PromptConf)}}~\citep{yoon2025reasoning} employs a prompt-based method, enabling LRMs to generate confidence scores during the reasoning process based on specific rules. At the end of the process, these scores are mapped into one of ten bins, ranging from “Almost no chance (0–0.1)” to “Almost certain (0.9–1.0).” Each bin includes both a linguistic descriptor (e.g., “Almost certain”) and its corresponding numerical probability $C_i$ (e.g., “0.9–1.0”) to improve interpretability. The decision to terminate reasoning is then made as follows:
\begin{equation}
    C_i = \pi_\theta(Q, T_{<i},\text{[Confidence Prompt]})    
\end{equation}
\begin{equation}
    s_i = \mathbb{I}(C_i > \lambda)
    \label{eq4}
\end{equation}
where $\pi_{\theta}$ represents the LRM parameterized by $\theta$, $\lambda$ is the confidence threshold, and $\mathbb{I}(\cdot)$ is an indicator function that returns true when $C_i$ exceeds $\lambda$. The method sequentially evaluates each intermediate answer in the reasoning trace, using the prompt to output confidence scores for the temporary answers.

\textbf{\textsc{Dynasor-CoT}}~\citep{fu2025reasoning} leverages a "Probe-In-The-Middle" approach, which appends carefully designed guidance prompts at intermediate stages of the reasoning process to explicitly elicit the model’s current answer (e.g., "Oh, I suddenly got the answer to the whole problem, Final Answer: \textbackslash boxed\{\}"). This method monitors LRMs at regular intervals (e.g., every 32, 64, or 128 tokens) to check for consistent answers across multiple intervals. Once consistent answers are detected, the reasoning process is terminated.

\subsection{Internal States-based}
\textbf{\textsc{Probe Confidence (ProbeConf)}}~\citep{zhang2025reasoning} trains a MLP-based probing model to detect the intermediate answer correctness. Specifically, the probing model takes the last layer hidden states at the last token position of $T_i$ as input $h_i$, output a probability $C_i$ of intermediate correctness by:
\begin{equation}
    C_i=\text{MLP}(h_i)
\end{equation}
where the decision to exit is made by comparing the obtained confidence with the empirical threshold $\lambda$, as described in Equation~\ref{eq4}.

\textbf{\textsc{DEER}}~\citep{yang2025dynamic} monitors the action transition points as potential early exit points (e.g. wait) to inducing intermediate answer, which incorporated the answer delimiters  (i.e., \textbackslash boxed\{\}) into the prompt to facilitate a more precise identification of the trial answers, as follows: $A_i=\text{LRM}(Q,T_{<i},I)$ where $Q$ denotes the input prompt, $T_{<i}$ denotes already generated thoughts, $I$ denotes the answer induced prompt and $A_i=[a_{0,i},a_{1,i},\cdots,a_{l,i}]$ is the trail answer.

The confidence evaluator module computes the confidence of the induced trial answer. We take the maximum predicted probability of each token as its confidence. For multi-token trial answers, the overall confidence is computed as the mean confidence across all constituent tokens as follows:
\begin{equation}
    p(a_{t,i})=\mathrm{softmax}(\mathcal{M}(Q,T_{<i},I,a_{<t,i})),\quad\mathcal{C}_i=\left(\prod_{j=1}^l\max_{a_{t,i}\in\mathcal{V}}p(a_{t,i})\right)^{1/l}
\end{equation}
where the $\mathcal{M}$ is the LM head of LRMs. Finally, the decision to exit early is made by comparing the obtained confidence $C_i$ with the empirical threshold $\lambda$, as described in Equation~\ref{eq4}.

\textbf{\textsc{Entropy}}~\citep{yong2025think} proves that each reasoning step contributes to reducing entropy over the answer space and increasing confidence in the correct answer. \textsc{Entropy} uses the similar trick in \textsc{DEER} to obtain the trail answer $A_i$ and compute the conditional entropy at step $i$:
\begin{equation}
    C_{i} = -\sum^{l}_{t=1} P(a_{t,i}\mid Q,T_{<i},I,a_{<t,i})\text{log}P(a_{t,i}\mid Q,T_{<i},I,a_{<t,i})
\end{equation}
where $P(a_{t,i}\mid Q,T_{<i},I,a_{<t,i})$ is estimated from the model’s output probabilities. After each intermediate reasoning step, the model computes the average entropy $C^{\text{avg}}_i=\frac{1}{l}\sum^l_{i=1}C_i$ over the answer space. Reasoning is terminated early once the average entropy falls below a confidence threshold, which is parameterized by a hyperparameter $\alpha\in[0,1]$:
\begin{equation}
    s_i = \mathbb{I} (C^{\text{avg}}_i \leq \alpha \cdot \frac{1}{e\text{ln}2})
\end{equation}

\section{Mode Selection As a Harder Early Exit Problem}
\label{sec:mode_selection}

As introduced in Section~\ref{sec:early_exit}, we construct a unified view of early exit. In this section, we first introduce the \textsc{NoThinking} mode as an alternative option for mode selection~\citep{ma2025reasoning} representing short-CoT. Recent studies have highlighted \textsc{NoThinking} as an effective method for adaptive reasoning, which can significantly reduce token usage while preserving performance~\citep{liang2025thinkswitcher, li2025dynamicmind}.
\subsection{NoThinking Mode}

\textsc{NoThinking} is designed to bypass the explicit reasoning process of LRMs by carefully crafting input prompts that already include the end-of-thinking delimiter \texttt{</think>} within pre-defined fake thoughts. This methodology has been adopted by systems like Qwen3~\citep{yang2025qwen3} and DeepSeek-V3.1~\citep{deepseekv3.1} to enable hybrid inference in LRMs. By using fake thoughts, many LRMs are able to save token usage while maintaining strong performance on simpler tasks. The process follows this pattern:

\begin{equation}
    \textsc{NoThinking}:  \underbrace{[\text{Prompt}] + \texttt{<think>} + \text{[Fake Thoughts]} + \texttt{</think>}}_{\text{Input}} + \underbrace{ \text{[Conclusion]}}_{\text{Generate}}
\end{equation}
where the \text{[Fake Thoughts]} for $T^{\text{fake}}_0$ fall into two categories: (i) an empty block, represented as \texttt{<think></think>}, and (ii) predefined thinking-completion statements (e.g., \texttt{<think>} Okay, I think I have finished thinking. \texttt{</think>}), which are intended to encourage LRMs to skip the reasoning process. We refer to this as \textit{Zero-Step Thinking}.
\subsection{Mode Selection formulates as Early Exit Problem}
In this paper, we present an interesting perspective: Mode Selection can be viewed as a more challenging variant of Early Exit for performing adaptive reasoning. As illustrated in Figure~\ref{fig:intro}, the key difference between these two approaches lies in their decision-making processes. Early Exit decides when to stop reasoning dynamically during the reasoning process, whereas Mode Selection operates as a static method, making decisions before explicit reasoning begins at the zero-step.

For simplicity, Early Exit iteratively performs $\text{Exit}(Q, T_{<i})$ at each reasoning step, while Mode Selection executes $\text{Exit}(Q, T_0^{\text{fake}})$ by replacing dynamically generated [Thoughts] with pre-defined [Fake Thoughts]. This means that Mode Selection lacks any question-specific reasoning information when making decisions, making the task significantly more difficult.
In this study, we conduct a systematic empirical investigation to explore whether reasoning models can effectively "think" using only zero-step thoughts.

\section{Experiments}
\label{sec:exp}
\subsection{Experiments Setup}
\paragraph{\textbf{Datasets.}}
We evaluate model performance across 4 benchmarks, including three mathematical reasoning benchmarks: GSM8K~\citep{cobbe2021training}, MATH-500~\citep{hendrycks2021measuring}, and AIME 2025~\citep{AIME}, as well as one scientific reasoning benchmark: GPQA Diamond~\citep{rein2024gpqa}. Among the mathematical reasoning benchmarks, GSM8K and MATH-500 are generally regarded as relatively simple reasoning tasks, whereas AIME 2025 is considered more challenging. 

\paragraph{\textbf{Metrics.}}
We selected Accuracy (\textbf{Acc}), Token Number (\textbf{Tok}), and \textsc{NoThinking} Ratio (\textbf{NR}) as the evaluation metrics. Acc denotes the final answer accuracy. Tok denotes the average generation length per sample to evaluate the cost.  NR denotes the choose of \textsc{NoThinking} mode ratio.

\paragraph{\textbf{Backbone LRMs.}}

We conducted experiments using the open-source DeepSeek-R1-Distill-Qwen series of models (1.5B, 7B, and 32B), which are distilled from DeepSeek-R1~\citep{guo2025deepseek}. These models utilize Qwen2.5-1.5B~\citep{yang2025qwen3} as the backbone and are fine-tuned on 800k high-quality reasoning samples. This fine-tuning process enables the models to achieve superior performance on logical and mathematical reasoning tasks.

\paragraph{\textbf{Baselines.}}
In this work, we aim to systematically analyze whether it is possible to use zero-step thoughts to perform Mode Selection. We evaluate six baseline methods discussed in Section~\ref{sec:early_exit}, including \textsc{FlashThink}, \textsc{PromptConf}, \textsc{Dynasor-CoT}, \textsc{ProbeConf}, \textsc{DEER}, and \textsc{Entropy}. Additionally, we introduce three new baselines: \textsc{Thinking}, \textsc{NoThinking}, and \textsc{Pre-judge}~\citep{zeyularge}. \textsc{Thinking} directly evaluates the performance of LRMs without any external intervention. \textsc{NoThinking} incorporates pre-defined fake thoughts to bypass explicit reasoning. \textsc{Pre-judge} prompts the LRMs to decide whether reasoning is required, based on carefully designed input prompts.
For all baselines, we first randomly sample instances for both \textsc{Thinking} and \textsc{NoThinking} with a temperature setting of 0.6. Next, we use the existing baselines to determine the flag $s_i$ for each instance. In the main results, we manually select the best performance for each baseline by varying the threshold $\lambda$. Detailed implementation details for each baseline are provided in Appendix~\ref{app:imple}.

{
\begin{table*}[]
\centering
\caption{Experimental results across various types of reasoning models are presented. "Acc" denotes accuracy, "Tok" represents the token count, and "NR" refers to the \textsc{NoThinking} mode rate. $\uparrow$ indicates that higher values are better, while $\downarrow$ indicates that lower values are preferable. Additionally, the baselines are categorized as follows: \colorbox{blue!10}{\textcolor{blue!10}{1}} represents basic baselines, \colorbox{yellow!20}{\textcolor{yellow!20}{1}} represents prompt-based methods, and \colorbox{red!20}{\textcolor{red!20}{1}} indicates internal states-based methods.}
\scalebox{0.52}{
\begin{tabular}{@{}lrrrrrrrrrrrr@{}} 
\toprule
 \multirow{2}{*}{\textbf{Method}} 
 & \multicolumn{3}{|c}{\textbf{GSM8K}}& \multicolumn{3}{c}{\textbf{MATH-500}}  & \multicolumn{3}{c}{\textbf{AIME25}}  & \multicolumn{3}{c}{\textbf{GPQA-D}}   \\
   &  \multicolumn{1}{c}{Acc$\uparrow$} &  \multicolumn{1}{c}{Tok$\downarrow$} &  \multicolumn{1}{c}{NR}  &  \multicolumn{1}{c}{Acc$\uparrow$} &  \multicolumn{1}{c}{Tok$\downarrow$} &  \multicolumn{1}{c}{NR} &  \multicolumn{1}{c}{{Acc}$\uparrow$}  &  \multicolumn{1}{c}{{Tok$\downarrow$}} &  \multicolumn{1}{c}{NR} &  \multicolumn{1}{c}{Acc$\uparrow$} &  \multicolumn{1}{c}{Tok$\downarrow$} &  \multicolumn{1}{c}{NR}  \\ 
\hline

\multicolumn{13}{l}{{\cellcolor[rgb]{0.957,0.957,0.957}}\textit{\textbf{DeepSeek-R1-Distill-Qwen-1.5B}}} \\
\colorbox{blue!10}{\textsc{Thinking}} & 85.7 \small{\textcolor{black}{(\textbf{0.0})}} & 2,455 \small{\textcolor{black}{(\textbf{0.0\%})}} & 0\% & 83.8 \small{\textcolor{black}{(-\textbf{0.0})}}& 5,445 \small{\textcolor{black}{(\textbf{0.0\%})}} & 0\%  &33.3 \small{\textcolor{black}{(-\textbf{0.0})}}& 15,214 \small{\textcolor{black}{(\textbf{0.0\%})}} & 0\% & 24.8 \small{\textcolor{black}{(-\textbf{0.0})}}& 9,818 \small{\textcolor{black}{(\textbf{0.0\%})}} & 0\%   \\

\colorbox{blue!10}{\textsc{NoThinking}} & 72.7 \small{\textcolor{teal}{(-\textbf{13.0})}} & 261 \small{\textcolor{blue}{(-\textbf{89\%})}} & 100\% & 68.6 \small{\textcolor{teal}{(-\textbf{15.2})}} & 1,411 \small{\textcolor{blue}{(-\textbf{74.1\%})}}& 0\% & 20.0 \small{\textcolor{teal}{(-\textbf{13.3})}} & 6,614 \small{\textcolor{blue}{(-\textbf{56.5\%})}}& 100\% & 18.2 \small{\textcolor{teal}{(-\textbf{6.6})}}& 945 \small{\textcolor{blue}{(-\textbf{90.4\%})}}& 100\%   \\

\colorbox{yellow!20}{\textsc{FlashThink}} & 85.7 \small{\textcolor{black}{(\textbf{0.0})}} & 2,455 \small{\textcolor{black}{(\textbf{0.0\%})}} & 100\% & 83.8 \small{\textcolor{black}{(-\textbf{0.0})}}& 5,445 \small{\textcolor{black}{(\textbf{0.0\%})}} &100\%  &33.3 \small{\textcolor{black}{(-\textbf{0.0})}}& 15,214 \small{\textcolor{black}{(\textbf{0.0\%})}} & 100\% & 24.8 \small{\textcolor{black}{(-\textbf{0.0})}}& 9,818 \small{\textcolor{black}{(\textbf{0.0\%})}} & 100\% \\  

\colorbox{yellow!20}{\textsc{PromptConf}}  & 85.0 \small{\textcolor{teal}{(-\textbf{0.7})}} & 2,272 \small{\textcolor{blue}{(-\textbf{7.5\%})}}& 6.6\%& 82.0 \small{\textcolor{teal}{(-\textbf{1.8})}} & 5,159 \small{\textcolor{blue}{(-\textbf{5.3\%})}}& 8.0\%  & 40.0 \small{\textcolor{red}{(+\textbf{6.7})}} & 9,731 \small{\textcolor{blue}{(-\textbf{36.0\%})}}& 66.7\% & 24.2 \small{\textcolor{teal}{(-\textbf{0.6})}} & 8,598 \small{\textcolor{blue}{(-\textbf{12.4\%})}}& 16.2\% \\

\colorbox{yellow!20}{\textsc{Dynasor-CoT}}  & 84.2 \small{\textcolor{teal}{(-\textbf{1.5})}}& 2,103 \small{\textcolor{blue}{(-\textbf{14.3\%})}} & 20.3\% & 80.8 \small{\textcolor{teal}{(-\textbf{3.0})}} & 4,752 \small{\textcolor{blue}{(-\textbf{12.7\%})}}& 23.2\% & 33.3  \small{\textcolor{black}{(-\textbf{0.0})}}& 14,227 \small{\textcolor{blue}{(-\textbf{6.5\%})}}& 6.7\% & 23.7 \small{\textcolor{teal}{(-\textbf{1.1})}}& 8,959 \small{\textcolor{blue}{(-\textbf{8.7\%})}}& 15.2\%\\

\colorbox{yellow!20}{\textsc{Pre-Judge}} & 77.8\small{\textcolor{teal}{(-\textbf{7.9})}} & 1,269 \small{\textcolor{blue}{(-\textbf{48.3\%})}}& 59.3\% & 81.8 \small{\textcolor{teal}{(-\textbf{2.0})}} & 5,100 \small{\textcolor{blue}{(-\textbf{6.3\%})}}& 12.0\% &  33.3 \small{\textcolor{black}{(-\textbf{0.0})}} & 15,214 \small{\textcolor{black}{(\textbf{0.0\%})}}& 100\% & 22.7 \small{\textcolor{teal}{(-\textbf{2.1})}} & 5,193 \small{\textcolor{blue}{(-\textbf{47.1\%})}}& 55.6\% \\  

\colorbox{red!20}{\textsc{ProbeConf}}  & 84.4\small{\textcolor{teal}{(-\textbf{1.3})}} & 2,299 \small{\textcolor{blue}{(-\textbf{6.4\%})}}& 9.2\% & 82.6 \small{\textcolor{teal}{(-\textbf{1.2})}} & 4,941 \small{\textcolor{blue}{(-\textbf{9.3\%})}}& 13.8\% & 40.0 \small{\textcolor{red}{(+\textbf{6.7})}}& 11,596 \small{\textcolor{blue}{(-\textbf{23.8\%})}}& 20.0\% & 22.7 \small{\textcolor{teal}{(-\textbf{2.1})}}& 7,399 \small{\textcolor{blue}{(-\textbf{24.6\%})}}& 26.8\% \\  

\colorbox{red!20}{\textsc{DEER}}  &     85.3 \small{\textcolor{teal}{(-\textbf{0.4})}}& 2,276 \small{\textcolor{blue}{(-\textbf{7.3\%})}} & 9.7\%&  82.6 \small{\textcolor{teal}{(-\textbf{1.2})}}& 5,067  \small{\textcolor{blue}{(-\textbf{6.9\%})}}& 11.6\% &  33.3 \small{\textcolor{black}{(-\textbf{0.0})}} & 8334 \small{\textcolor{blue}{(-\textbf{45.2\%})}}& 53.3\%   & 25.3 \small{\textcolor{red}{(+\textbf{0.5})}}& 9,274 \small{\textcolor{blue}{(-\textbf{5.5\%})}}& 9.6\% \\

\colorbox{red!20}{\textsc{Entropy} } &  85.3 \small{\textcolor{teal}{(-\textbf{0.4})}} & 2,278 \small{\textcolor{blue}{(-\textbf{7.2\%})}}&8.5\%& 82.6 \small{\textcolor{teal}{(-\textbf{1.2})}}& 5,199 \small{\textcolor{blue}{(-\textbf{4.5\%})}}& 11.0\%  & 40.0  \small{\textcolor{red}{(+\textbf{6.7})}}   & 12,784 \small{\textcolor{blue}{(-\textbf{16.0\%})}}& 26.7\%  & 28.3 \small{\textcolor{red}{(+\textbf{3.5})}}& 6,268 \small{\textcolor{blue}{(-\textbf{36.2\%})}}& 43.4\%  \\

\hline

\multicolumn{13}{l}{{\cellcolor[rgb]{0.957,0.957,0.957}}\textit{\textbf{DeepSeek-R1-Distill-Qwen-7B}}} \\
\colorbox{blue!10}{\textsc{Thinking}} & 92.3 \small{\textcolor{black}{(-\textbf{0.0})}} & 1,687 \small{\textcolor{black}{(-\textbf{0.0\%})}} & 0\% & 93.0 \small{\textcolor{black}{(-\textbf{0.0})}} & 4,100 \small{\textcolor{black}{(-\textbf{0.0\%})}} & 0\%  & 40.0 \small{\textcolor{black}{(-\textbf{0.0})}}& 15,024 \small{\textcolor{black}{(-\textbf{0.0\%})}} & 0\% & 47.5 \small{\textcolor{black}{(-\textbf{0.0})}} & 8,447 \small{\textcolor{black}{(-\textbf{0.0\%})}}& 0\%   \\

\colorbox{blue!10}{\textsc{NoThinking}} & 87.6 \small{\textcolor{teal}{(-\textbf{4.7})}} & 268 \small{\textcolor{blue}{(-\textbf{84.1\%})}} & 100\% & 78.0 \small{\textcolor{teal}{(-\textbf{15.0})}}& 781 \small{\textcolor{blue}{(-\textbf{81.0\%})}}& 100\% & 26.7 \small{\textcolor{teal}{(-\textbf{13.3})}}& 1,352 \small{\textcolor{blue}{(-\textbf{91.0\%})}}& 100\% & 19.2 \small{\textcolor{teal}{(-\textbf{28.3})}}& 688 \small{\textcolor{blue}{(-\textbf{91.9\%})}}& 100\%   \\

\colorbox{yellow!20}{\textsc{FlashThink}} & 92.3 \small{\textcolor{black}{(-\textbf{0.0})}} & 1,687 \small{\textcolor{black}{(-\textbf{0.0\%})}} & 0\% & 93.0 \small{\textcolor{black}{(-\textbf{0.0})}} & 4,100 \small{\textcolor{black}{(-\textbf{0.0\%})}} & 0\%  & 40.0 \small{\textcolor{black}{(-\textbf{0.0})}}& 15,024 \small{\textcolor{black}{(-\textbf{0.0\%})}} & 0\% & 47.5 \small{\textcolor{black}{(-\textbf{0.0})}} & 8,447 \small{\textcolor{black}{(-\textbf{0.0\%})}}& 0\%\\  

\colorbox{yellow!20}{\textsc{PromptConf}}  & 89.5 \small{\textcolor{teal}{(-\textbf{2.8})}} & 691 \small{\textcolor{blue}{(-\textbf{59.0\%})}}& 72.9\%& 82.6 \small{\textcolor{teal}{(-\textbf{10.4})}}& 1,776 \small{\textcolor{blue}{(-\textbf{56.7\%})}}& 76.8\%  & 33.3 \small{\textcolor{teal}{(-\textbf{6.7})}}& 8,301 \small{\textcolor{blue}{(-\textbf{44.7\%})}}& 60.0\% & 29.3 \small{\textcolor{teal}{(-\textbf{18.2})}}& 4,093 
\small{\textcolor{blue}{(-\textbf{51.5\%})}}& 54.5\% \\

\colorbox{yellow!20}{\textsc{Dynasor-CoT}}  & 92.1 \small{\textcolor{teal}{(-\textbf{0.2})}}& 1330 \small{\textcolor{blue}{(-\textbf{21.2\%})}} & 30.0\% & 89.2 \small{\textcolor{teal}{(-\textbf{3.8})}}& 2,993 \small{\textcolor{blue}{(-\textbf{27.0\%})}}& 34.6\% & 40.0 \small{\textcolor{black}{(-\textbf{0.0})}} & 13,587 \small{\textcolor{blue}{(-\textbf{9.6\%})}}& 13.3\% & 31.3 \small{\textcolor{teal}{(-\textbf{16.2})}}& 4778 \small{\textcolor{blue}{(-\textbf{43.4\%})}}& 53.0\%\\

\colorbox{yellow!20}{\textsc{Pre-Judge}} & 91.3\small{\textcolor{teal}{(-\textbf{1.0})}} & 1,211 \small{\textcolor{blue}{(-\textbf{28.2\%})}}& 37.1\% & 90.0 \small{\textcolor{teal}{(-\textbf{3.0})}}& 3,381 \small{\textcolor{blue}{(-\textbf{17.5\%})}}& 26.6\% & 40.0 \small{\textcolor{black}{(-\textbf{0.0})}}& 14,875 \small{\textcolor{blue}{(-\textbf{1.0\%})}}& 6.7\% & 38.9 \small{\textcolor{teal}{(-\textbf{8.6})}}& 6,925 \small{\textcolor{blue}{(-\textbf{18.0\%})}}& 20.7\% \\  

\colorbox{red!20}{\textsc{ProbeConf}}  & 91.6\small{\textcolor{teal}{(-\textbf{0.7})}} & 1,394 \small{\textcolor{blue}{(-\textbf{17.4\%})}}& 18.1\% & 92.2 \small{\textcolor{teal}{(-\textbf{0.8})}}& 3,950 \small{\textcolor{blue}{(-\textbf{27.5\%})}}& 8.4\% & 46.7 \small{\textcolor{red}{(+\textbf{6.7})}}& 11,025 \small{\textcolor{blue}{(-\textbf{26.6\%})}}& 26.7\% & 47.0 \small{\textcolor{teal}{(-\textbf{0.5})}}& 8,190 \small{\textcolor{blue}{(-\textbf{3.0\%})}}& 6.6\% \\  

\colorbox{red!20}{\textsc{DEER}}       & 92.6 \small{\textcolor{red}{(+\textbf{0.3})}}& 1,476 \small{\textcolor{blue}{(-\textbf{10.8\%})}} & 18.1\%&  93.2 \small{\textcolor{red}{(+\textbf{0.2})}}& 3,912  \small{\textcolor{blue}{(-\textbf{4.6\%})}}& 8.4\% &  40.0 \small{\textcolor{black}{(-\textbf{0.0})}}& 14,603 \small{\textcolor{blue}{(-\textbf{2.8\%})}}& 6.7\%   & 47.0 \small{\textcolor{teal}{(-\textbf{0.5})}}& 8,269 \small{\textcolor{blue}{(-\textbf{2.1\%})}}& 4.0\% \\

\colorbox{red!20}{\textsc{Entropy} } &   92.2\small{\textcolor{teal}{(-\textbf{0.1})}} & 1,505 \small{\textcolor{blue}{(-\textbf{7.2\%})}}& 18.1\%& 92.6 \small{\textcolor{teal}{(-\textbf{0.4})}}& 3,871 \small{\textcolor{blue}{(-\textbf{5.6\%})}}& 8.4\%  & 40.0  \small{\textcolor{black}{(-\textbf{0.0})}}& 12,784 \small{\textcolor{blue}{(-\textbf{14.9\%})}}& 26.7\%  & 48.0 \small{\textcolor{red}{(+\textbf{0.5})}}& 6,693 \small{\textcolor{blue}{(-\textbf{20.8\%})}}& 26.8\%  \\

\hline

\multicolumn{13}{l}{{\cellcolor[rgb]{0.957,0.957,0.957}}\textit{\textbf{DeepSeek-R1-Distill-Qwen-32B}}} \\
\colorbox{blue!10}{\textsc{Thinking}} & 95.8 \small{\textcolor{black}{(-\textbf{0.0})}} & 1,453 \small{\textcolor{black}{(-\textbf{0.0\%})}} & 0\% & 94.0 \small{\textcolor{black}{(-\textbf{0.0})}} & 3,462 \small{\textcolor{black}{(-\textbf{0.0\%})}} & 0\%  & 66.7 \small{\textcolor{black}{(-\textbf{0.0})}}& 11,155 \small{\textcolor{black}{(-\textbf{0.0\%})}} & 0\% & 62.1 \small{\textcolor{black}{(-\textbf{0.0})}} & 6,690 \small{\textcolor{black}{(-\textbf{0.0\%})}}& 0\%   \\

\colorbox{blue!10}{\textsc{NoThinking}} & 95.9 \small{\textcolor{red}{(+\textbf{0.1})}} & 1,280 \small{\textcolor{blue}{(-\textbf{11.9\%})}} & 100\% & 94.2 \small{\textcolor{red}{(+\textbf{0.2})}}& 3,550 \small{\textcolor{magenta}{(+\textbf{2.5\%})}}& 100\% & 60.0 \small{\textcolor{teal}{(-\textbf{6.7})}}& 13,933 \small{\textcolor{magenta}{(+\textbf{24.9\%})}}& 100\% & 62.6 \small{\textcolor{red}{(+\textbf{0.5})}}& 6,576 \small{\textcolor{blue}{(-\textbf{1.7\%})}}& 100\%   \\

\colorbox{yellow!20}{\textsc{FlashThink}} &  95.8 \small{\textcolor{black}{(-\textbf{0.0})}} & 1,453 \small{\textcolor{black}{(-\textbf{0.0\%})}} & 0\% & 94.0 \small{\textcolor{black}{(-\textbf{0.0})}} & 3,462 \small{\textcolor{black}{(-\textbf{0.0\%})}} & 0\%  & 66.7 \small{\textcolor{black}{(-\textbf{0.0})}}& 11,155 \small{\textcolor{black}{(-\textbf{0.0\%})}} & 0\% & 62.1 \small{\textcolor{black}{(-\textbf{0.0})}} & 6,690 \small{\textcolor{black}{(-\textbf{0.0\%})}}& 0\% \\  

\colorbox{yellow!20}{\textsc{PromptConf}}  & 95.6 \small{\textcolor{teal}{(-\textbf{0.2})}} & 1,338 \small{\textcolor{blue}{(-\textbf{59.0\%})}}& 83.1\%& 94.0 \small{\textcolor{black}{(-\textbf{0.0})}}& 3,488 \small{\textcolor{magenta}{(+\textbf{0.8\%})}}& 64.2\%  & 60.0 \small{\textcolor{teal}{(-\textbf{6.7})}}& 11,295 \small{\textcolor{magenta}{(+\textbf{1.3\%})}}& 26.7\% & 61.6 \small{\textcolor{teal}{(-\textbf{0.5})}}& 6,557 
\small{\textcolor{blue}{(-\textbf{2.0\%})}}& 31.3\% \\

\colorbox{yellow!20}{\textsc{Dynasor-CoT}}  & 95.7 \small{\textcolor{teal}{(-\textbf{0.1})}}& 1,384 \small{\textcolor{blue}{(-\textbf{7.9\%})}} & 33.7\% & 93.8 \small{\textcolor{teal}{(-\textbf{0.2})}}& 3,392 \small{\textcolor{blue}{(-\textbf{2.0\%})}}& 39.0\% & 66.7 \small{\textcolor{black}{(-\textbf{0.0})}} & 11,054 \small{\textcolor{blue}{(-\textbf{0.9\%})}}& 13.3\% & 63.6 \small{\textcolor{red}{(+\textbf{1.5})}}& 6,699 \small{\textcolor{magenta}{(+\textbf{0.1\%})}}& 46.5\%\\

\colorbox{yellow!20}{\textsc{Pre-Judge}} & 95.7\small{\textcolor{teal}{(-\textbf{0.1})}} & 1,317 \small{\textcolor{blue}{(-\textbf{9.4\%})}}& 37.1\% & 94.4 \small{\textcolor{red}{(+\textbf{0.4})}}& 3,467 \small{\textcolor{magenta}{(+\textbf{0.1\%})}}& 39.4\% & 66.7 \small{\textcolor{black}{(-\textbf{0.0})}}& 11,155 \small{\textcolor{black}{(-\textbf{0.0\%})}} & 0\% & 62.1 \small{\textcolor{black}{(-\textbf{0.0})}}& 6,669 \small{\textcolor{blue}{(-\textbf{0.3\%})}}& 0.51\% \\  

\colorbox{red!20}{\textsc{ProbeConf}}  & 96.0\small{\textcolor{red}{(+\textbf{0.2})}} & 1,389 \small{\textcolor{blue}{(-\textbf{4.4\%})}}& 38.1\% & 94.2 \small{\textcolor{red}{(+\textbf{0.2})}}& 3,413 \small{\textcolor{blue}{(-\textbf{1.4\%})}}& 20.8\% & 66.7 \small{\textcolor{black}{(-\textbf{0.0})}}& 11,182 \small{\textcolor{magenta}{(+\textbf{0.2\%})}}& 20.0\% & 65.2 \small{\textcolor{red}{(+\textbf{3.1})}}& 6,312 \small{\textcolor{blue}{(-\textbf{5.7\%})}}& 90.4\% \\  

\colorbox{red!20}{\textsc{DEER}}       & 96.1 \small{\textcolor{red}{(+\textbf{0.3})}}& 1,342 \small{\textcolor{blue}{(-\textbf{7.6\%})}} &  53.4\%&  94.2 \small{\textcolor{red}{(+\textbf{0.2})}}& 3,419  \small{\textcolor{blue}{(-\textbf{1.2\%})}}& 20.8\% &  66.7 \small{\textcolor{black}{(-\textbf{0.0})}}& 10,947 \small{\textcolor{blue}{(-\textbf{1.9\%})}}& 53.3\%   & 65.7 \small{\textcolor{red}{(+\textbf{3.6})}}& 6,686 \small{\textcolor{blue}{(-\textbf{0.1\%})}}& 73.7\% \\

\colorbox{red!20}{\textsc{Entropy} } &   95.7\small{\textcolor{teal}{(-\textbf{0.1})}} & 1,423 \small{\textcolor{blue}{(-\textbf{2.1\%})}}& 38,1\%& 94.2 \small{\textcolor{red}{(+\textbf{0.2})}}& 3,512 \small{\textcolor{magenta}{(+\textbf{1.4\%})}}& 40.8\%  & 66.7  \small{\textcolor{black}{(-\textbf{0.0})}}& 10,567 \small{\textcolor{blue}{(-\textbf{5.3\%})}}& 46.7\%  & 62.6 \small{\textcolor{red}{(+\textbf{0.5})}}& 6578 \small{\textcolor{blue}{(-\textbf{1.7\%})}}& 90.4\%  \\
 \bottomrule
\end{tabular}
}
\label{tab:main}
\end{table*}
}

\subsection{Main Results}

In this section, we present the main experimental results and an in-depth analysis of how different types of baselines perform on the Mode Selection task.
Notably, different baselines exhibit varying decision strategies. For \textsc{FlashThink} and \textsc{Pre-Judge}, the model determines the mode at the zero-step stage without relying on threshold-based decisions. In contrast, baselines like \textsc{PromptConf} and \textsc{Dynasor-CoT}, which depend on fixed discrete scores, consistently use the highest score as the threshold $\lambda$. However, for baselines such as \textsc{DEER}, \textsc{ProbeConf}, and \textsc{Entropy}, fixed thresholds fail to effectively capture their optimal performance. Consequently, we manually selected the best-performing thresholds for these baselines in the main experiment.
\paragraph{\textbf{Limitations of Prompt-based Methods.}} We begin by analyzing prompt-based methods, which rely on a verification model to decide whether to terminate the reasoning process. Due to the limited information available from fake thoughts $T^\text{fake}_0$, \textsc{FlashThink} consistently determines that LRMs must continue reasoning, leading to a 0\% NR rate across all scenarios. Moreover, the performance of other methods varies significantly across different datasets and base models. For example, \textsc{PromptConf} achieves notable token reductions while maintaining performance stability on smaller models (e.g., a 6.7 accuracy improvement on AIME25 with a 36.0\% reduction in token usage for a 1.5B model). However, its effectiveness decreases as model size increases (e.g., 7B and 32B models). Similar trends are observed in \textsc{Dynasor-CoT} and \textsc{PromptConf}. Nonetheless, these two methods demonstrate improved stability by leveraging self-generated scores to evaluate reasoning states.
\paragraph{\textbf{Internal States Tell More Than Language.}}
Due to the limited effectiveness of prompt-based methods, we further explore approaches that leverage model internal states, including \textsc{ProbeConf}, \textsc{DEER}, and \textsc{Entropy}. For manual selection of $\lambda$, we first retain thresholds that yield significant accuracy improvements. If such improvements are absent, we instead select thresholds that align the accuracy or NR across these three methods, which facilitates more consistent analysis. As shown in Table~\ref{tab:main}, on the 1.5B model, \textsc{DEER} and \textsc{Entropy} achieve better accuracy retention on GSM8k and GPQA-D, while \textsc{ProbeConf} provides higher token compression in MATH-500 and even yields a 6.7-point accuracy gain on AIME25 with 26.6\% token compression. A similar trend is observed for the 7B and 32B models. These methods consistently reduce token usage while maintaining performance, and in some cases even surpass baseline results. For example, \textsc{DEER} achieves superior performance on the 32B model, reducing token usage while preserving accuracy and even outperforming \textsc{Thinking}. Overall, the results demonstrate that signals from the model’s internal states provide more reliable indicators for selecting the appropriate mode.

\begin{figure}[]
    \centering
    \begin{subfigure}{0.32\textwidth}
        \centering
        \includegraphics[width=\textwidth]{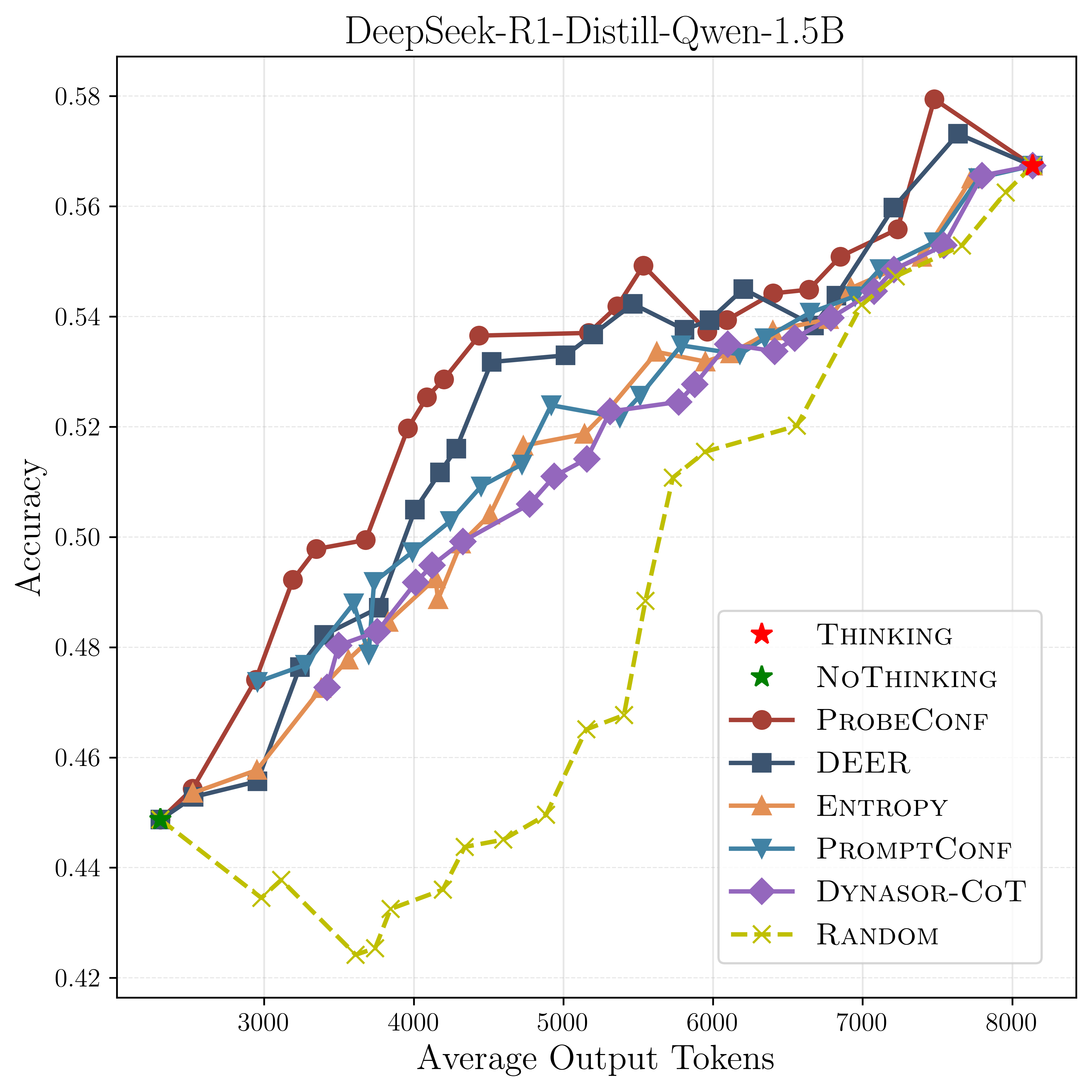}
    \end{subfigure}
    \hfill
    \begin{subfigure}{0.32\textwidth}
        \centering
        \includegraphics[width=\textwidth]{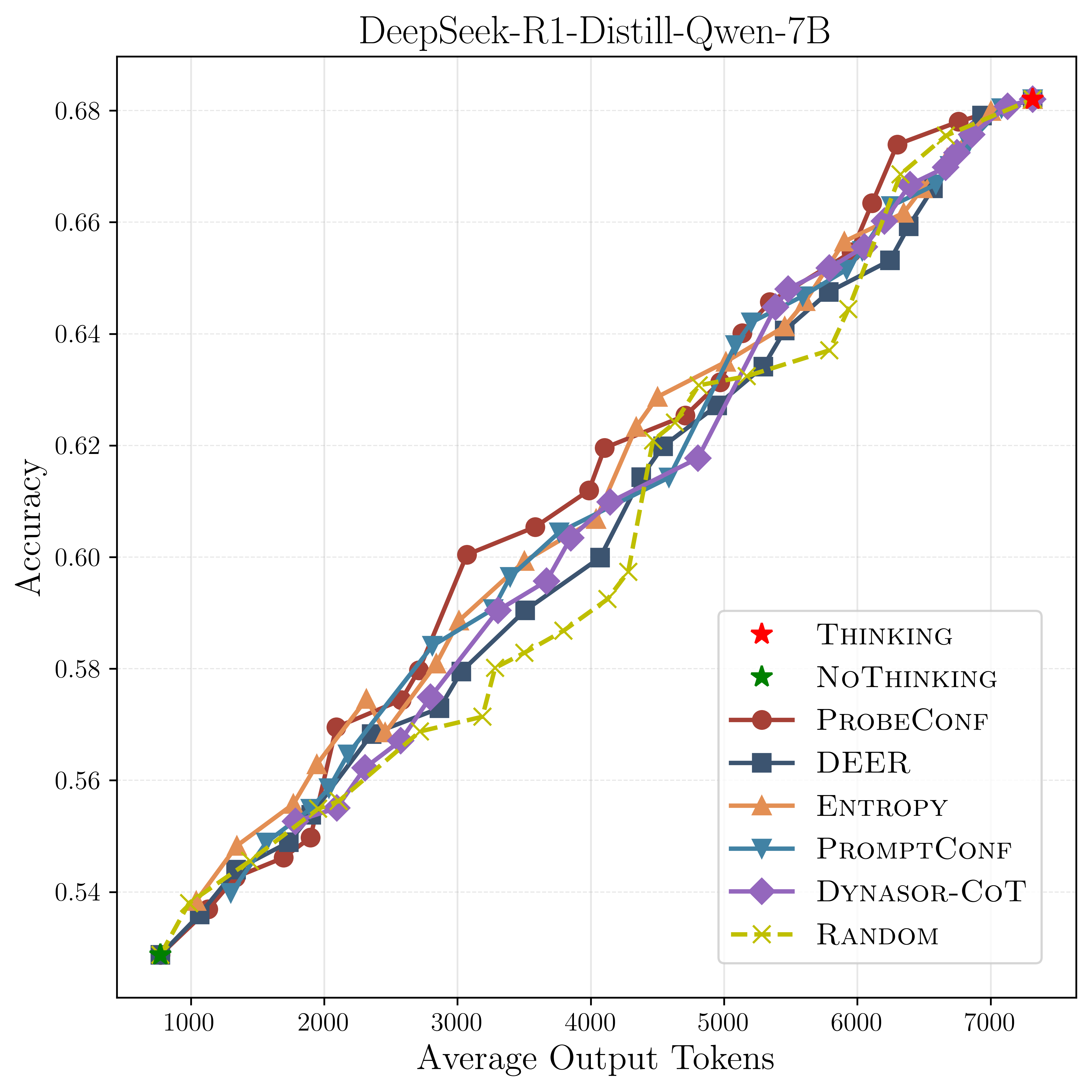}
    \end{subfigure}
    \hfill
    \begin{subfigure}{0.32\textwidth}
        \centering
        \includegraphics[width=\textwidth]{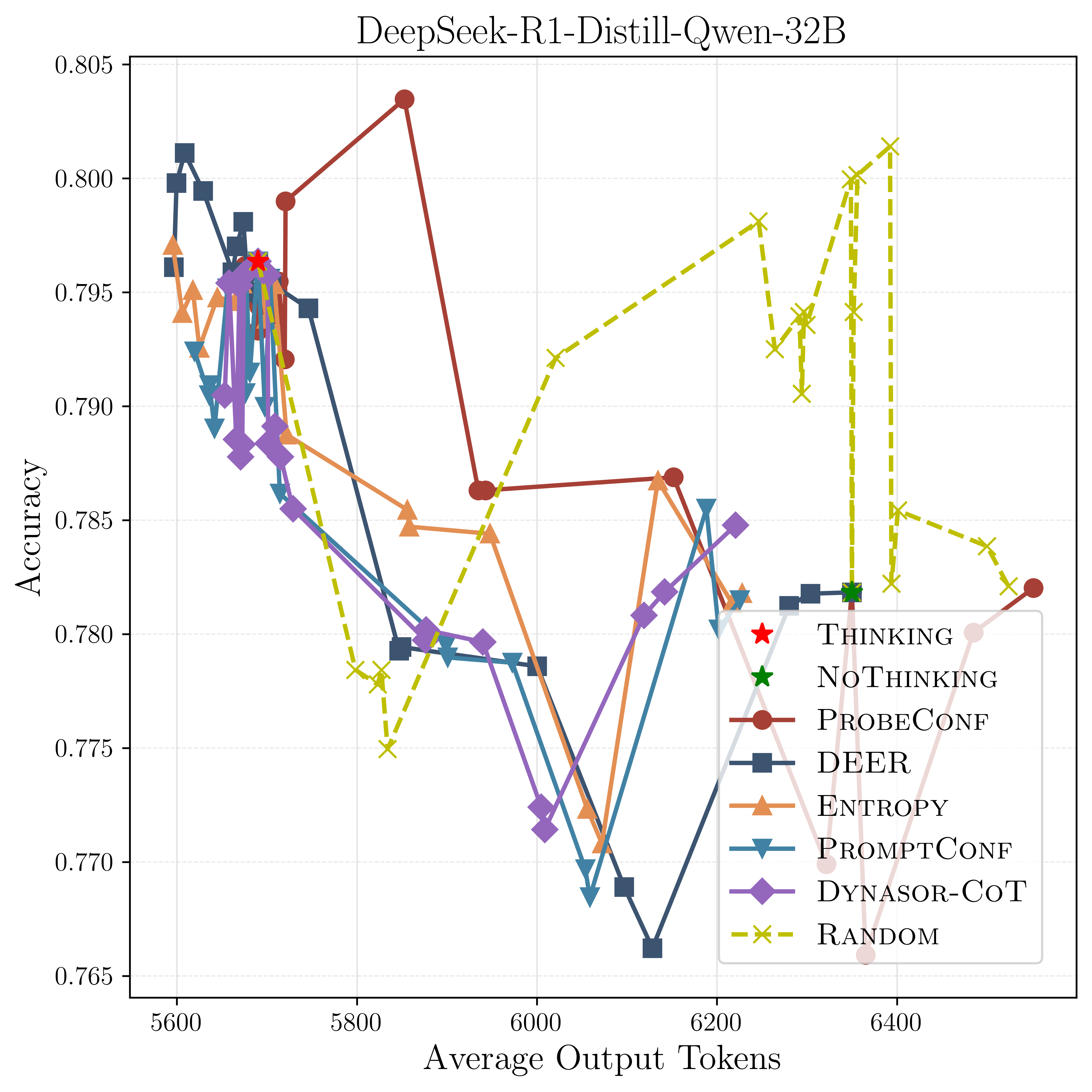}
    \end{subfigure}
    \hfill
    \caption{Trade-off between average accuracy and average output tokens across different methods and three model sizes. Each point on the curve represents a distinct $\lambda$ value from 0.1 to 1.0.}
    \label{fig:lambda}
\end{figure}
\section{Analyses}
To better understand the behavior and effectiveness of these methods in the Mode Selection task, we conduct detailed analyses. These include examining threshold dynamics (Section~\ref{sec:threshold_dyna}), evaluating ROC-AUC scores across methods (Section~\ref{sec:ROC-AUC}), and analyzing the correlation between $C_i$ and \textsc{NoThinking} mode accuracy (Section~\ref{sec:ece}).

\subsection{Analysis of Threshold Dynamics} 
\label{sec:threshold_dyna}
Since internal state–based methods primarily rely on a well-defined threshold $\lambda$, whose optimal value varies unpredictably across tasks and models, we evaluate the trade-off between accuracy and cost across three model scales: DeepSeek-R1-Distill-Qwen-1.5B, 7B, and 32B. For each model, we plot average accuracy against average token cost by systematically sweeping the decision threshold $\lambda$. The resulting trade-off curves are shown in Figure~\ref{fig:lambda}.

On the 1.5B model, all methods consistently outperform the random baseline, demonstrating their ability to dynamically select appropriate reasoning modes to balance performance and computational cost. In particular, \textsc{ProbeConf} and \textsc{DEER} show stronger performance than the other methods, suggesting that leveraging internal states can yield a better Pareto frontier by achieving more favorable accuracy–cost trade-offs across operating points.

A clear trend emerges in the accuracy–cost curves under varying thresholds. For the 1.5B model, mode selection methods are clearly distinguished from the random baseline, highlighting their effectiveness on smaller models. However, as the scale increases to 7B, the performance gap between methods narrows. For example, even the best-performing method, \textsc{ProbeConf}, occasionally underperforms relative to the \textsc{Random} baseline, resulting in weaker separation. This phenomenon becomes more pronounced with the 32B model, where the effectiveness of \textsc{Thinking} and \textsc{NoThinking} is reversed (Table~\ref{tab:main}). On datasets such as MATH-500 and AIME25, \textsc{NoThinking} generates more tokens than \textsc{Thinking}, because certain examples under the \textsc{NoThinking} strategy still produce lengthy outputs. We hypothesize that this behavior arises because LRMs have internalized the reasoning process: forcing them to bypass it with fake thoughts not only fails to elicit direct summarization but may also cause the model to restart its reasoning. A similar trend has also been observed in QwQ-32B by~\citep{zhu2025can}.
Therefore, in the following sections, we focus on the 1.5B and 7B models to gain deeper insights into the factors influencing performance in mode selection.

\begin{figure}[t]
    \centering
    
    \begin{subfigure}{1\textwidth}
        \centering
        \includegraphics[width=1\textwidth]{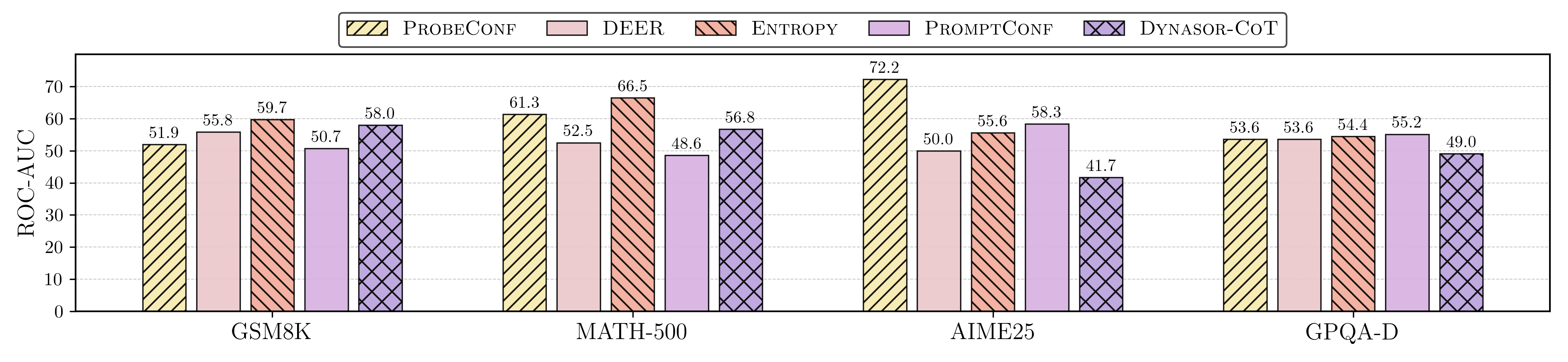}
        \caption{ROC AUC comparison for DeepSeek-R1-Distill-Qwen-1.5B model.}
        \label{fig:roc_1.5b}
    \end{subfigure}

    \begin{subfigure}{1\textwidth}
        \centering
        \includegraphics[width=1\textwidth]{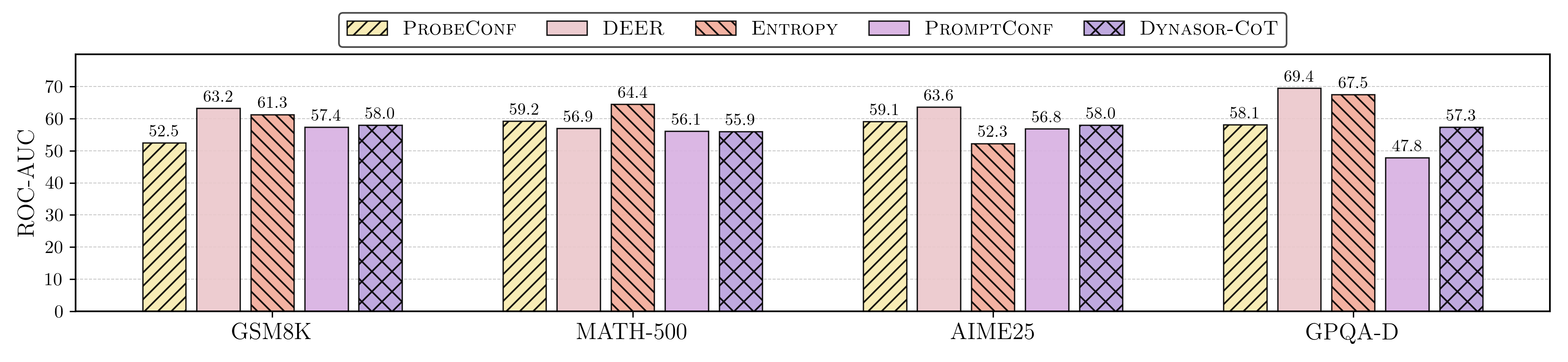}
        \caption{ROC AUC comparison for DeepSeek-R1-Distill-Qwen-7B model.}
        \label{fig:roc_7b}
    \end{subfigure}
    
    \caption{ROC-AUC comparison between different model sizes: (a) 1.5B and (b) 7B.}
    \label{fig:roc_comparison}
\end{figure}

\subsection{Analysis of ROC-AUC across methods}
\label{sec:ROC-AUC}
In this section, we move beyond fixed thresholds and instead use ROC-AUC scores for more robust comparisons. As shown in Figure~\ref{fig:roc_comparison}, the results reinforce our earlier findings: these methods are highly sensitive to both dataset and model, and no single approach consistently achieves optimal performance across all scenarios. For example, on the 1.5B model, \textsc{ProbeConf} attains the highest ROC-AUC score of 72.2 on AIME25, substantially outperforming other methods. However, this advantage does not generalize to the 7B model or to other datasets. Overall, methods leveraging internal states tend to outperform prompt-based approaches in most cases, suggesting that a model’s intrinsic states encode richer information than the lossy representations conveyed by language.

\begin{table}[!t]
 \caption{Expected Calibration Error (ECE) and Brier score in 1.5B and 7B cross datasets. $\downarrow$ means smaller is better.
    }
    \label{tab:calib}
    
    \centering
    \scalebox{0.88}{
    \begin{tabular}{lcccccccc}
            \toprule
            \multirow{2}{*}{\textbf{Baselines}} & \multicolumn{2}{c}{\textbf{GSM8K}} & \multicolumn{2}{c}{\textbf{MATH-500}} & \multicolumn{2}{c}{\textbf{AIME25}} & \multicolumn{2}{c}{\textbf{GPQA-D}} \\
            \cmidrule{2-9}
            & \textbf{ECE $\downarrow$} & \textbf{Brier $\downarrow$} & \textbf{ECE $\downarrow$} & \textbf{Brier $\downarrow$} & \textbf{ECE $\downarrow$} & \textbf{Brier $\downarrow$} & \textbf{ECE $\downarrow$} & \textbf{Brier $\downarrow$} \\
            \midrule
            \multicolumn{9}{l}{{\cellcolor[rgb]{0.957,0.957,0.957}}\textit{\textbf{DeepSeek-R1-Distill-Qwen-1.5B}}} \\
\textsc{ProbeConf} &    \textbf{0.122} & \textbf{0.215} & \textbf{0.078} & \textbf{0.215} & 0.447  & 0.320  &   0.407 & 0.322         \\
\textsc{deer} &  0.218 & 0.262 & 0.198 & 0.270 & \textbf{0.224} & \textbf{0.203} & \textbf{0.297} & \textbf{0.250} \\

\textsc{dynasor-cot} &    0.235 & 0.261         &    0.188 & 0.264  & 0.200 & 0.259     & 0.441 & 0.391    \\

            \multicolumn{9}{l}{{\cellcolor[rgb]{0.957,0.957,0.957}}\textit{\textbf{DeepSeek-R1-Distill-Qwen-7B}}} \\
\textsc{ProbeConf} &    0.296 & 0.222 & \textbf{0.128} & \textbf{0.190} & 0.401  & 0.364 & 0.528  & 0.436              \\
\textsc{deer} & 0.272 & \textbf{0.213} & 0.192 & 0.232 & 0.145 & \textbf{0.181} & \textbf{0.354} & \textbf{0.273} \\

\textsc{dynasor-cot} &    \textbf{0.268} & \textbf{0.213}         &    0.229 & 0.234  & \textbf{0.111} & 0.200    & 0.636 & 0.579       \\
            \bottomrule
        \end{tabular}}
   
\end{table}

\subsection{Correlation Analysis of $C_i$ with \textsc{NoThinking} Mode Accuracy}
\label{sec:ece}
Since these methods rely on the \textsc{NoThinking} prompt template to compute a score $C_i$ as the mode selection indicator, we further investigate whether $C_i$ correlates with answer correctness in \textsc{NoThinking} mode. To this end, we report the corresponding Expected Calibration Error (ECE)~\citep{naeini2015obtaining} and Brier score~\citep{glenn1950verification}. For a fair comparison, \textsc{Entropy} is excluded due to numerical issues.

As shown in Table~\ref{tab:calib}, different methods exhibit varying trends across datasets. Notably, \textsc{ProbeConf} and \textsc{DEER} generally yield lower errors compared to the prompt-based method \textsc{Dynasor-CoT}. Moreover, when examining how these metrics align with performance, we observe that the Brier score on the 7B model is highly consistent with the ROC-AUC results in Figure~\ref{fig:roc_comparison}. For example, \textsc{DEER} achieves the lowest Brier score of 0.181 on AIME25, corresponding to the best ROC-AUC among all methods. In contrast, these metrics appear less informative for the 1.5B model. We hypothesize that smaller models suffer a substantial drop in performance under \textsc{NoThinking} mode, making both internal and external scores less reliable for performance estimation.

\section{Related Work}
\paragraph{\textbf{Adaptive Thinking for Overthinking.}} Recent studies have shown that longer CoT do not always improve performance~\citep{chen2024not, wu2025more}, and in some cases may even lead to overthinking, particularly in high-capacity models~\citep{li2025thinkless}. This has sparked growing interest in minimal or implicit reasoning strategies~\citep{ma2025reasoning}, highlighting the need for more nuanced approaches and adaptive control of reasoning depth through RL or related techniques~\citep{fang2025thinkless, yang2025towards, tan2025dynamic}. In this work, we focus on a specific subtype of adaptive thinking, namely mode selection, which aims to determine the appropriate reasoning mode prior to explicit reasoning. Existing approaches either train a routing mechanism or directly prompt the model to decide~\citep{liang2025thinkswitcher, fu2025reasoning, tan-etal-2025-neural}. In contrast, we present a systematic analysis of how methods developed for Early Exit can be leveraged to address this challenging problem.

\paragraph{\textbf{Early Exit for Efficient Reasoning.}}
Efficiency in LLMs is an active research area, with methods that adapt the number of reasoning steps according to task difficulty, confidence, or resource constraints~\citep{shen2025dast, li2025dynamicmind}. Among these methods, Early Exit has proven especially effective: it determines an optimal stopping point, reducing token usage and sometimes even improving performance. Formally, this process can be abstracted as the function $\text{Exit}(Q, T_{<i})$~\citep{yang2025dynamic, jiang2025flashthink, zhang2025reasoning}.
Building on how such methods are implemented, we focus on prompt-based approaches, which rely on the reasoning model itself or an auxiliary model to decide whether to stop thinking based on textual information~\citep{jiang2025flashthink, yoon2025reasoning, fu2025reasoning}. In contrast, internal state–based approaches leverage intrinsic signals from the model, such as hidden states or output logits, to make this decision~\citep{yang2025dynamic, zhang2025reasoning, yong2025think}. We extend this framework to the Mode Selection paradigm, where the task becomes more complex as performing $\text{Exit}(Q, T^{\text{fake}}_0)$.

\section{Conclusion}
In this work, we introduce Mode Selection as a more challenging variant of the Early Exit problem, where the model should determine the appropriate reasoning mode before explicit reasoning begins. Through extensive empirical studies on multiple reasoning benchmarks, we find that prompt-based approaches are often limited by weak classification capability under minimal information, while internal states–based approaches achieve better performance but suffer from instability. Our analysis further shows that existing evaluation metrics are insufficient to fully explain method behaviors, highlighting the complexity of Mode Selection. Overall, our findings point to the need for more robust approaches that better exploit model internal mechanisms of \textsc{Thinking} and \textsc{NoThinking}, paving the way for future research on adaptive reasoning strategies in LRMs.
\section*{Acknowledgments}
This work was supported by the National Key R\&D Program of China (No. 2022ZD0160503) and Beijing Natural Science Foundation (L243006) and the National Natural Science Foundation of China (No.62376270).

\newpage
\bibliography{neurips_2025}
\bibliographystyle{plain}


\appendix
\section{Implementation Details}
\label{app:imple}
\subsection{Basic Setup}
\paragraph{\textbf{Decoding details.}}
All evaluations were conducted in a zero-shot Chain-of-Thought (CoT) setting using the prompt: "Please reason step by step, and put your final answer within \textbackslash boxed{}." For decoding, we used sampling with a temperature of 0.6. The ground-truth answers in our experiments consist exclusively of well-structured numerical values or categorical options; accordingly, we applied rule-based checks to verify mathematical equivalence. We set the maximum generation length to 16,384 tokens to ensure complete problem-solving attempts were captured.

\paragraph{\textbf{Benchmarks.}}To comprehensively evaluate the models’ reasoning capabilities, we employ four representative benchmarks widely used in the field. GSM8K~\citep{cobbe2021training} is a carefully curated dataset of 1,319 elementary mathematics problems, specifically designed to assess multi-step reasoning in foundational math tasks. Each problem typically requires two to eight sequential operations, relying primarily on basic arithmetic applied across multiple intermediate steps. MATH-500~\citep{hendrycks2021measuring} is a challenging benchmark composed of competition-level problems drawn from diverse high school mathematics domains, including Prealgebra, Algebra, and Number Theory. For consistency with prior research, we adopt the 500-problem subset originally curated by OpenAI.
AIME 2025~\citep{AIME} consists of 30 problems selected from the 2025 American Invitational Mathematics Examination (AIME). This prestigious contest evaluates mathematical reasoning across a broad range of domains, including arithmetic, algebra, counting, geometry, number theory, probability, and other advanced secondary school topics.
Beyond math, we also evaluate on scientific reasoning tasks. GPQA~\citep{rein2024gpqa} is a PhD-level benchmark covering physics, chemistry, and biology. Domain experts with PhDs in these areas achieve only 69.7\% accuracy on this dataset~\citep{rein2024gpqa}, highlighting its difficulty. For our experiments, we specifically use the highest-quality subset, GPQA Diamond, which comprises 198 questions.

\subsection{Baselines Implementations}
In this section, we provide the clear implementation of each baseline methods.

\paragraph{\textbf{\textsc{Thinking}}.} The \textsc{Thinking} mode refers to the default chat template of existing LRMs~\citep{guo2025deepseek, yang2025qwen3} which only appends the \texttt{<think>} token to enable thinking process:
\begin{tcolorbox}[colback=gray!20, colframe=gray!50!black, title=Prompt Template for \textsc{Thinking}]
<BOS\_TOKEN><|USER|>\{Question\} \\
Please reason step by step, and put your final answer within \textbackslash boxed\{\}. \\
<|Assistant|><think>
\end{tcolorbox}

\paragraph{\textbf{\textsc{NoThinking}}.} The \textsc{NoThinking} mode proposed by~\citep{ma2025reasoning} can significant reduce token usage to achieve better performance for sampling more times. In recent studies, it has been seen as a straight but useful baseline to stimulate short-CoT ability of LRMs by appending fake thoughts with \texttt{</think>} to skip the thinking process. We use the same prompt template as follow:
\begin{tcolorbox}[colback=gray!20, colframe=gray!50!black, title=Prompt Template for \textsc{NoThinking}]
<BOS\_TOKEN><|USER|>\{Question\}

Please reason step by step, and put your final answer within \textbackslash boxed\{\}.

<|Assistant|><think>

Okay, I think I have finished thinking.

</think>
\end{tcolorbox}

\paragraph{\textbf{\textsc{FlashThink}}.} \textsc{FlashThink} utilizes another verification model $\pi_{\phi}$ to decide whether to skip thinking. Following~\citep{jiang2025flashthink}, we select Qwen2.5-7B-Instruct\footnote{https://huggingface.co/Qwen/Qwen2.5-7B-Instruct} as $\pi_\phi$ to verify with the following prompt where the \{Question\} will be replace with the real question $Q$:

\begin{tcolorbox}[colback=gray!20, colframe=gray!50!black, title=Prompt Template for \textsc{FlashThink}]
<BOS\_TOKEN><|USER|>\\
Based on the following question and thought, please judge whether the thought is sufficient to support solving the question. Please directly output yes or no instead of outputting other content.\\
\#\#\# Question\\
\{Question\}\\
\#\#\# Thought\\
Okay, I think I have finished thinking.\\
<|Assistant|><think> 
\end{tcolorbox}

\paragraph{\textbf{\textsc{PromptConf}}.} \textsc{PromptConf} utilizes prompt method to make LRMs themselves generate confidence score during thinking process based on specific rule. The scores range from “Almost no chance (0–0.1)” to “Almost certain (0.9–1.0)”, and we dirrectly append "0." to make LRMs output correct pattern and we select the lower bound of the output range as final score:
\begin{tcolorbox}[colback=gray!20, colframe=gray!50!black, title=Prompt Template for \textsc{PromptConf}]
<BOS\_TOKEN><|USER|>

For the following question, classify your confidence into one of the following classes based on how likely your answer is to be correct:  \\
- "Almost no chance" (0.0-0.1) \\
- "Highly unlikely" (0.1-0.2) \\
- "Chances are slight" (0.2-0.3) \\
- "Unlikely" (0.3-0.4) \\
- "Less than even" (0.4-0.5) \\
- "Better than even" (0.5-0.6) \\
- "Likely" (0.6-0.7) \\
- "Very good chance" (0.7-0.8) \\
- "Highly likely" (0.8-0.9) \\
- "Almost certain" (0.9-1.0) \\
Each category reflects the probability that your answer is correct.\\
At the end of your output, format your answer and confidence as\\
Confidence: \$SCORE \\
where SCORE is one of the probability ranges of the scores above.\\
Here is the question:\\
\{Question\}

<|Assistant|><think> \\
</think>\\
Confidence: 0.
\end{tcolorbox}

\paragraph{\textbf{\textsc{Dynasor-CoT}}.} \textsc{Dynasor-CoT} employs a carefully designed guidance prompt to elicit the model’s immediate answer (e.g., "Oh, I suddenly got the answer to the whole problem, Final Answer: \textbackslash boxed\{..."). However, mode selection is a static process and thus cannot adopt the monitoring setup in~\citep{yoon2025reasoning}, which checks intermediate states at regular intervals. Inspired by self-consistency~\citep{wang2022self}, we randomly sample three outputs and use the maximum agreement ratio as the score. This results in three possible scores 33.3\%, 66.7\%, and 100\%, where higher values indicate greater model confidence on a given question.
\begin{tcolorbox}[colback=gray!20, colframe=gray!50!black, title=Prompt Template for \textsc{Dynasor-CoT}]
<BOS\_TOKEN><|USER|>\{Question\}

Please reason step by step, and put your final answer within \textbackslash boxed\{\}.

<|Assistant|><think>

Okay, I think I have finished thinking.

Oh, I suddenly got the answer to the whole problem, Final Answer: \textbackslash boxed\{
\end{tcolorbox}

\paragraph{\textbf{\textsc{Pre-Judge}}.} \textsc{Pre-Judge} is similar to \textsc{FlashThink} but uses LRM itself to verify.  We use the same prompt in~\citep{zeyularge} to output the boolean value of 'require\_slow\_thinking' as $s_i$:

\begin{tcolorbox}[colback=gray!20, colframe=gray!50!black, title=Prompt Template for \textsc{Pre-Judge}]
<BOS\_TOKEN><|USER|>

You are a math problem solver. For the following question, determine if it requires slow thinking or can be solved quickly. You do not need to give me any explanation, just give me a json with the following keys: require\_slow\_thinking. \\
For example: \{'require\_slow\_thinking': true\}\\
Here is the question:\\
\{Question\}

<|Assistant|><think> \\
</think>\\
\{'require\_slow\_thinking':
\end{tcolorbox}

\paragraph{\textbf{\textsc{ProbeConf}}.} \textsc{ProbeConf} relys on a MLP-based probing model to detect the intermediate answer correctness. For the probing models we used in our experiments, we adopt the already trained models which trained in MATH-500\footnote{https://github.com/AngelaZZZ-611/reasoning\_models\_probing} , and using the same training script to perform grid search to obtain best hyper-parameters. During evaluation, we use the same prompt template of \textsc{NoThinking} to extract the hidden states $h_i$ of \texttt{</think>}:

\begin{tcolorbox}[colback=gray!20, colframe=gray!50!black, title=Prompt Template for \textsc{ProbeConf}]
<BOS\_TOKEN><|USER|>\{Question\}

Please reason step by step, and put your final answer within \textbackslash boxed\{\}.

<|Assistant|><think>

Okay, I think I have finished thinking.

</think>
\end{tcolorbox}

\paragraph{\textbf{\textsc{DEER}}.} \textsc{DEER} utilizes the induced prompt $I$ to output the intermediate answer, and calculate $C_i$ based on the average of output logits to show model's confidence: 
\begin{tcolorbox}[colback=gray!20, colframe=gray!50!black, title=Prompt Template for \textsc{DEER}]
<BOS\_TOKEN><|USER|>\{Question\}

Please reason step by step, and put your final answer within \textbackslash boxed\{\}.

<|Assistant|><think>

Okay, I think I have finished thinking.

</ think>

**Final Answer**

The final answer is \textbackslash boxed\{
\end{tcolorbox}

\paragraph{\textbf{\textsc{Entropy}}.} \textsc{Entropy} leverages the entropy of output logits to estimate the model’s confidence. For simplicity, we compute entropy only at the final thinking position to assess whether the LRM exhibits sufficient confidence:

\begin{tcolorbox}[colback=gray!20, colframe=gray!50!black, title=Prompt Template for \textsc{Entropy}]
<BOS\_TOKEN><|USER|>\{Question\}

Please reason step by step, and put your final answer within \textbackslash boxed\{\}.

<|Assistant|><think>

Okay, I think I have finished thinking.

\end{tcolorbox}

\end{document}